%% file: main.tex
\documentclass[10pt,twocolumn,letterpaper]{article}

\usepackage[submission]{iccv}      
\DeclareUnicodeCharacter{202F}{\,} 

\definecolor{iccvblue}{rgb}{0.21,0.49,0.74}
\usepackage[pagebackref,breaklinks,colorlinks,allcolors=iccvblue]{hyperref}
\usepackage{times}
\usepackage{epsfig}
\usepackage{graphicx}
\usepackage{amsmath}
\usepackage{amssymb,amsfonts}

\usepackage{xspace}
\usepackage[misc]{ifsym}
\usepackage{multicol}
\usepackage{stfloats}
\usepackage{makecell}
\usepackage{pifont}
\usepackage{colortbl}
\usepackage{tabularx}
\usepackage{caption}

\usepackage{textcomp}
\usepackage{url}
\usepackage{verbatim}
\usepackage{graphicx}
\usepackage{color}
\usepackage{booktabs}
\usepackage{multirow}
\usepackage{bbding}

\def\paperID{4807} 
\def\confName{ICCV}
\def\confYear{2025}

\title{MobileViCLIP: An Efficient Video-Text Model for Mobile Devices}

\author{Min Yang\textsuperscript{1}  \quad Zihan Jia\textsuperscript{1}  \quad Zhilin Dai\textsuperscript{1}  \quad Sheng Guo\textsuperscript{2}  \quad Limin Wang\textsuperscript{1,3,~\Letter}\\
$^1$State Key Laboratory for Novel Software Technology, Nanjing University  \\
\quad $^2$MyBank, Ant Group \quad $^3$Shanghai AI Lab \\
{\tt\small  yangminmcg1011@hotmail.com, zihanjia@smail.nju.edu.cn, daizhilin1@gmail.com, } \\
{\tt\small guosheng1001@gmail.com
, lmwang@nju.edu.cn} \\
}

\begin{document}
\maketitle

\begin{abstract}
\hspace{1em}
Efficient lightweight neural networks are with increasing attention due to their faster reasoning speed and easier deployment on mobile devices. However, existing video pre-trained models still focus on the common ViT architecture with high latency, and few works attempt to build efficient architecture on mobile devices. 
This paper bridges this gap by introducing temporal structural reparameterization into an efficient image-text model and training it on a large-scale high-quality video-text dataset, resulting in an efficient video-text model that can run on mobile devices with strong zero-shot classification and retrieval capabilities, termed as \textbf{MobileViCLIP}.
In particular, in terms of inference speed on mobile devices, our MobileViCLIP-Small is \textbf{55.4}$\times$ times faster than InternVideo2-L14 and \textbf{6.7}$\times$ faster than InternVideo2-S14. In terms of zero-shot retrieval performance, our MobileViCLIP-Small obtains similar performance as InternVideo2-L14 and obtains 6.9\% better than InternVideo2-S14 on MSR-VTT. The code is available at \textbf{\url{https://github.com/MCG-NJU/MobileViCLIP}}.

\end{abstract}
{
\renewcommand{\thefootnote}%
{\fnsymbol{footnote}}
\footnotetext[0]{\Letter~Corresponding author.} 
}
\section{Introduction}
\hspace{1em}
Large vision-text models~\cite{internvl,coca,blip2,omnivec} have demonstrated excellent zero-shot performance in a wide range of downstream tasks. Learning transferable video-text representations is becoming a fundamental task in video understanding. With the advent of high-quality video-language datasets~\cite{webvid,internvid,openvid,panda-70m} for large-scale pre-training, several powerful video-text foundation models~\cite{internvideo,internvideo2,unmaskedteacher} have emerged. They are typically trained from scratch and have achieved excellent results on various downstream tasks including video-text retrieval and action recognition. However, to ensure performance, most of these works use ViT-L~\cite{vit} or larger backbones. The model size limits its possibility of deployment on mobile devices with limited storage capability. Also, excessive computing resource requirements (tens of A100 GPUs are required), too long training time (thousands of hours of GPU training time), and a large amount of pre-training datasets (ensemble of datasets larger than 100M) increase the training cost and make these works difficult to reproduce for a university lab.

\begin{figure}[!t]
  \includegraphics[width=0.5\textwidth]{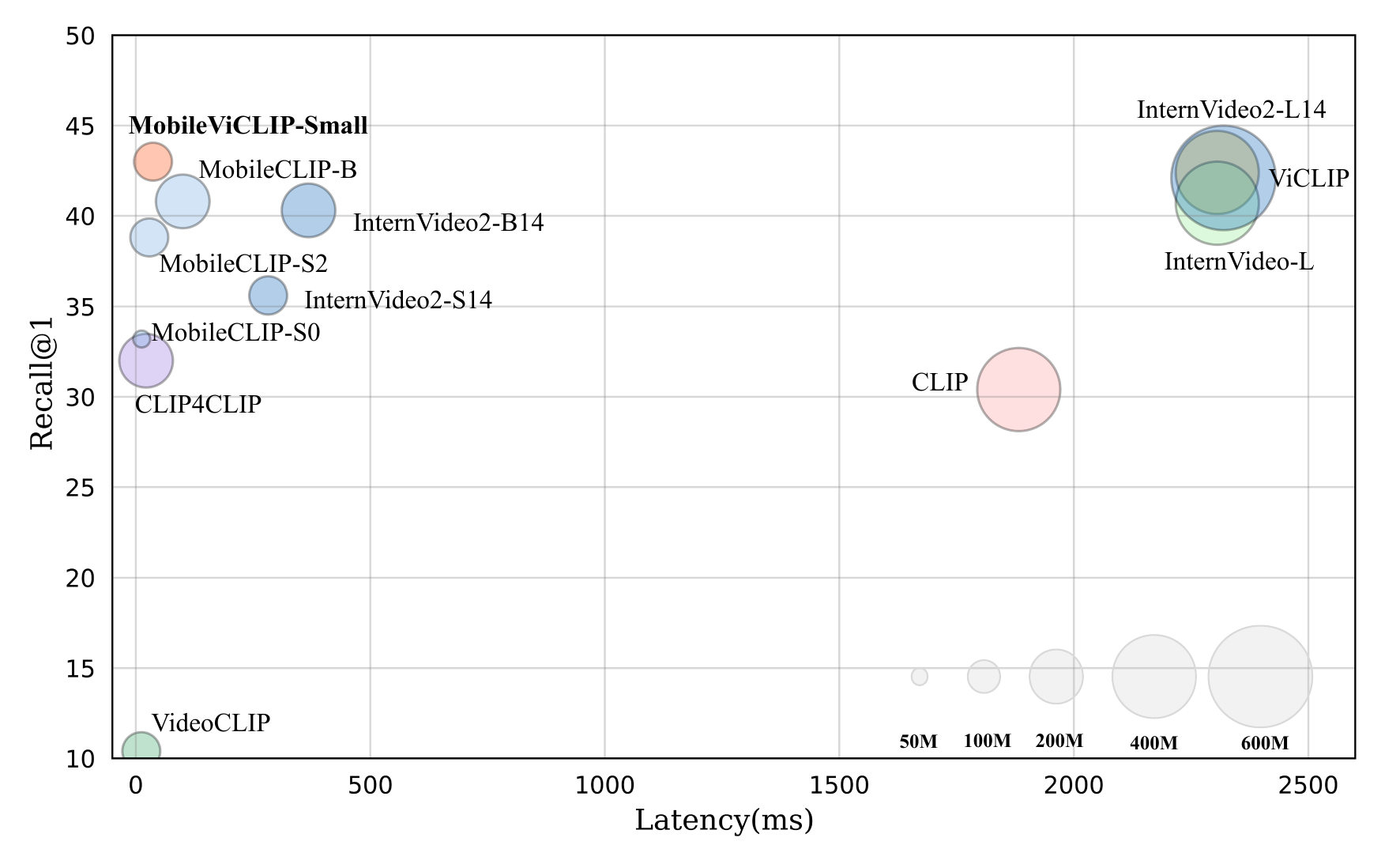}
  \caption{
\textbf{Comparison of latency and the R@1 scores for zero-shot text-to-video retrieval on video retrieval dataset MSR-VTT~\cite{msrvtt}.} The latency is measured on iPad Air 2020 with iOS 18. MobileViCLIP achieves the best trade-off of recall and latency.
}
\label{fig:intro}
\vspace{-3mm}
\end{figure}
Parameter-efficient transfer learning (PETL)~\cite{zeroi2v,st-adapter,aim,frozen} is becoming a promising and efficient paradigm for adapting large image-text models for video understanding due to their huge numbers of parameters and the high computational cost of full fine-tuning. These methods have successfully transferred image representations to video recognition tasks and obtained good zero-shot recognition results on several action classification datasets~\cite{k400,ucf101,hmdb51,ssv2}. 
However, these transfer learning methods often focus on the simple video classification task and fail to directly investigate the visual-textual representations on more complex video-text tasks. We argue that video-text models are expected to be more flexible and can handle a more diverse of downstream video understanding tasks. Furthermore, despite the extremely small number of trainable parameters in these transfer learning works, their used image-text models are still too large (ViT-B or ViT-L) to be deployed on mobile devices. It still remains a challenge that how to develop an efficient video-text model for mobile devices.

For image-text models, there have been several works~\cite {repnext,repvgg,repvit,mobileone,fastclip,fastvit} exploring efficient neural network architectures for mobile devices. Most of them adopt a hybrid architecture, by combining ViTs to capture global context through self-attention mechanisms and lightweight CNNs for their parameter efficiency and low latency. With the help of high-quality and large-scale image-text datasets~\cite{datacomp,laion-400m,openai-400m}, these models~\cite{slip,democratizingclip,mobileclip} have achieved good results on multiple image classification and retrieval tasks. However, the domain gap between image and video and the lack of temporal modeling limits their capabilities in video-text understanding. We aim to bridge this gap by adapting these image clip models to enable video understanding on mobile devices.

Specifically, we design an efficient video-text model that is able to run on a mobile device, coined as \textbf{MobileViCLIP}. 
The number of parameters in MobileViCLIP is smaller than all of these existing video-text models, which makes it possible to be deployed on mobile devices. In addition, the latency of MobileViCLIP is comparable with other efficient image-text models, which is significantly faster than the existing video-text models. 
To avoid the complex design and multi-stage training process of previous video-text models~\cite{internvideo2,vindlu}, we resort to the PETL paradigm by leveraging the image-text understanding capability of an existing model~\cite{mobileclip} and retraining on the additional high-quality and large-scale video-text datasets~\cite{internvid} to learn powerful and transferable video-text representations. To endow MobileViCLIP with the temporal modeling power, we replace the original block with Spatiotemporal RepMixer and Spatiotemporal Attention to capture both temporal dynamic and spatial representation. 
We only use \textbf{8 NVIDIA GeForce RTX 3090 GPUs} to train our MobileViCLIP by fully fine-tuning the video branch and freezing the text branch. It only takes \textbf{2 days} for training on the InternVid dataset. 
As shown in Fig~\ref{fig:intro}, the resulting MobileViCLIP-Small obtains comparable zero-shot text-to-video retrieval results to the ViT-L model from InternVideo2~\cite{internvideo2} (coined as InternVideo2-L14) with much lower latency and fewer parameters. 
In summary, our contributions are as follows:
\begin{itemize}
    \item
    We introduce \textbf{MobileViCLIP}, the \textbf{first} video-text model with efficient structure that is able to be efficiently deployed on mobile devices. Our MobileViCLIP has learned good video-text representations and achieved excellent zero-shot evaluation results on multiple datasets. 
    \item 
    We are the \textbf{first} to present an in-depth analysis on the latency of each basic module in current video-text models on the mobile devices, trying to figure out the key ingredients for future design of efficient video-text models. 
    \item 
    We verify the generalization performance of MobileViCLIP on downstream tasks including text-video retrieval and temporal grounding, demonstrating stronger generalization performance compared to the baseline. 
\end{itemize}

\section{Related Work}
\paragraph{\bf{Vision-Language Pretraining.}}  
By utilizing web-scale paired image-text data for training, CLIP~\cite{clip} shows robust zero-shot object recognition abilities. Also, some advanced works~\cite{eva-clip,internvl,eva-clip-18b,openclip} have explored the upper bound of scaling up contrastive language image pre-training. These models have billions of parameters, requiring billions of pre-training datasets and tens of thousands of hours of GPU training time. These works have achieved results comparable to fully supervised models on widely used recognition and retrieval datasets. The success of contrastive language-image pre-training has also inspired pre-training tasks for other modalities like video and audio.
\vspace{-5mm}
\paragraph{\bf{Video-Language Pretraining.}}  
The success of CLIP inspired the video domain. The study of video-language pretraining is becoming increasingly crucial considering its wide applications. Following the idea of image-text pre-training, several foundation models in the video domain have been born~\cite{cpd,videoclip,unmaskedteacher,vast,videoprism,violet,allinone}. These works explore the relationship between video modality and other modalities from different perspectives, achieving leading results across various video tasks. Undoubtedly, these works still require huge computing resources and training data, and the number of model parameters is huge. Our MobileViCLIP attempts to propose a foundation model that is mobile-friendly, with faster inference speed, lower latency, and fewer parameters within limited computing resources.
\vspace{-5mm}
\paragraph{\bf{CLIP-based Action Models.}}  
Inspired by the strong representation of the pre-trained image-text CLIP model, many video recognition approaches~\cite{actionclip,stan,vifi-clip,x-clip,adaptformer,st-adapter,aim,froster,zeroi2v} have been proposed based on CLIP. Although these works successfully transfer knowledge from image to video domain, these models are still far from the foundation models. These models are usually fully fine-tuned~\cite{actionclip,stan,vifi-clip} on video classification datasets~\cite{k400,ucf101,hmdb51}, or partially fine-tuned~\cite{adaptformer,st-adapter,x-clip,aim,zeroi2v} based on a proxy video action classification dataset. These models usually do not have video-text retrieval capability and cannot serve as feature extractors for downstream tasks. Our MobileViCLIP is fine-tuned based on a large-scale pre-training video-text dataset~\cite{internvid}, showing excellent generalization on more tasks. 

\vspace{-5mm}
\paragraph{\bf{Efficient CLIP Models.}}  
Recently, more and more works~\cite{mobileone,mobileclip,fastvit,repnext,repvgg,repvit} have realized the importance of deploying CLIP models on mobile devices, and propose a wide range of architectures that have shown great promise for accomplishing vision tasks on resource constraint devices. These architectures can be broadly classified
into purely convolutional~\cite{repvgg,mobilenetv3,mobileone}, transformer based~\cite{swintransformer,efficientvit} and convolution-transformer
hybrids like~\cite{fastvit,repnext,repvit}. However, to take into account low latency on mobile devices, the architectures of these models are generally smaller and their accuracy is much lower than that of conventional CLIP models~\cite{eva-clip}. To solve this problem, MobileCLIP tried to enhance the dataset~\cite{datacomp} with additional information and learned a good image-text representation. Based on it, we add efficient temporal modeling modules in Video TokenMixer and fine-tune it using a large-scale, high-quality video multimodal dataset~\cite{internvid}, so that MobileViCLIP can still demonstrate excellent zero-shot performance on multiple video-text retrieval and action recognition tasks while maintaining low latency.

\begin{figure}[!t]
  \includegraphics[width=0.5\textwidth]{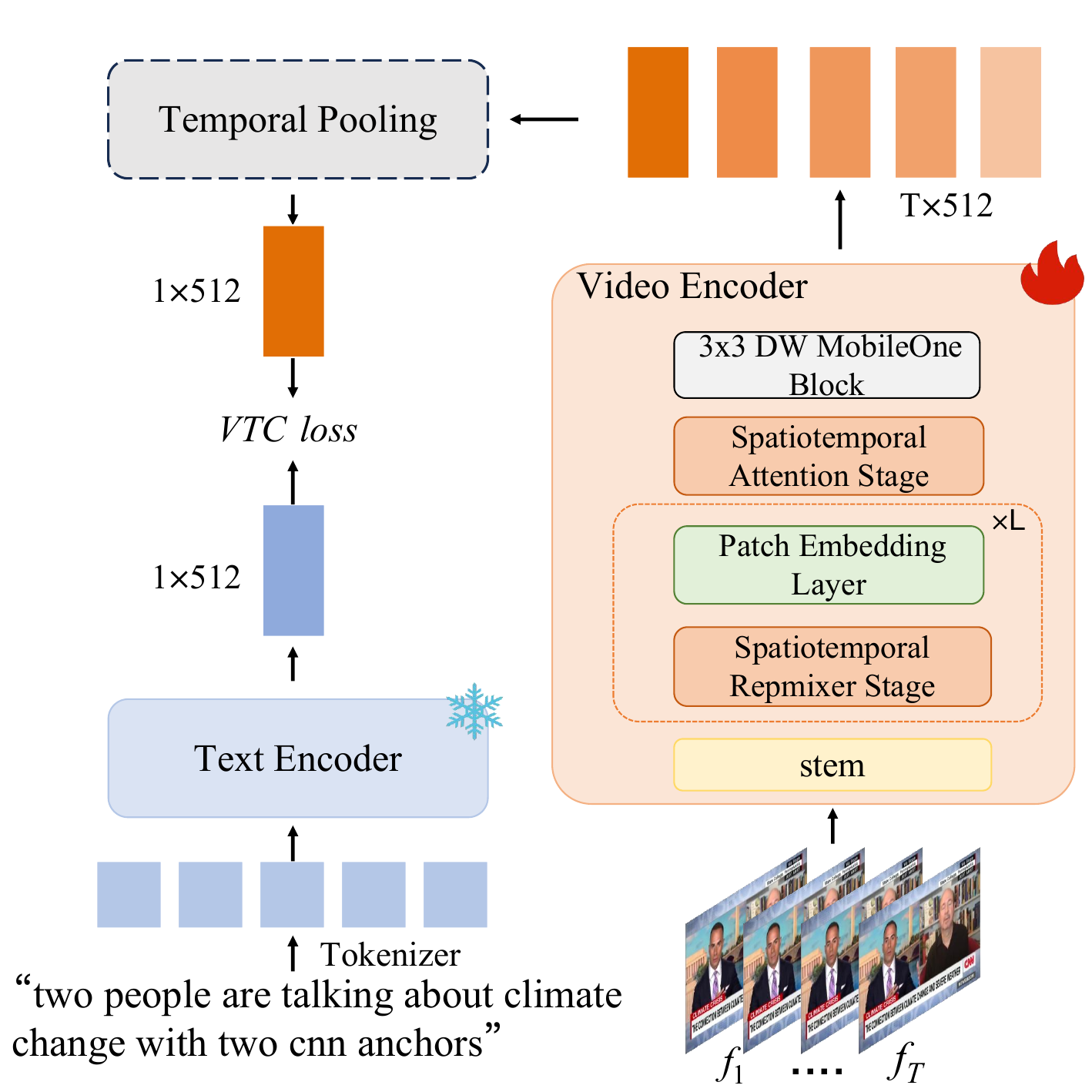}
  \caption{
\textbf{Overview of MobileViCLIP.} We fine-tune MobileCLIP~\cite{mobileclip} with minimal changes, which means we only introduce simple module modification to the Repmixer and Attention stage to convert the image encoder into a video encoder. During training, we freeze the text encoder and fully fine-tune the video encoder. We use a simple temporal pooling operation without any trainable parameters to generate the video-level representation for similarity matching with corresponding embeddings. 
}
\label{fig:overview}
\vspace{-4mm}
\end{figure}

\section{Method}

\subsection{MobileCLIP Revisited}
\paragraph{\bf{Backbone Baseline.}} 
We start with the description of image-text backbone designed by MobileCLIP~\cite{mobileclip}. We adopt two versions of MobileCLIP: MobileCLIP-S0 and MobileCLIP-S2, and build our MobileViCLIP-Tiny and MobileViCLIP-Small, respectively. For the image backbone, it is an improved hybrid vision transformer called MCi based on FastViT~\cite{fastvit}. MobileCLIP-S0 adopts MCi0 which has a similar stage configuration as MobileOne~\cite{mobileone} while MobileCLIP-S2 adopts MCi2 which is a wider and deeper version of MCi0.
For the text encoder, MobileCLIP-S0 uses MCt which is an efficient hybrid text encoder inspired by the reparameterizable convolutional token mixing (RepMixer) in FastViT~\cite{fastvit}. MobileCLIP-S2 adopts a 12-layer Transformer similar to that of ViT-B/16-CLIP~\cite{clip}.
\vspace{-4mm}
\paragraph{\bf{Structural Reparameterization.}} 
Recent works~\cite{fastvit,repvgg} show the benefits of reparameterizing skip connections to lower memory access costs. In our model, we follow the similar design in RepMixer, that is reparameterizable at inference. Specifically, for an input tensor $X$, the mixing block in the layer can be implemented as:
{\small
\[ Y = \mathrm{DWConv2D}(\mathrm{BN}(X)) + X.\]
}
And it can be reparameterized at inference time to a single depthwise convolutional layer without non-linear activation function $\sigma$ and Batch Normalization $BN$ as: 
{\small
\[Y = \mathrm{DWConv2D}(X). \]
}

\subsection{Overview of MobileViCLIP}
We describe the details of our MobileViCLIP pipeline shown in Fig~\ref{fig:overview}. Given a video sample $V \in R^{T \times H \times W \times 3}$ with T frames and corresponding text description, the video encoder encodes T frames and produces frame-level embeddings $f_{images} \in R^{T \times D}$. These frame-level embeddings are average-pooled to obtain a video-level representation $f_{video} \in R^{D}$ during inference. We refer to this as temporal pooling as this operation implicitly incorporates temporal learning via aggregation of multiple frames. The CLIP text encoder encodes the text descriptions to produce text embedding $t \in R^{D}$. Then we adopt video-text contrastive learning (VTC) to learn independent representations for video and text by maximizing the agreement between positive (video, text) pairs while minimizing the agreement between negative pairs. During inference, we only extract features through the video encoder and text encoder, and the pooling operation depends on the task requirements to choose whether to use. 

\begin{figure}[!t]
  \includegraphics[width=0.5\textwidth]{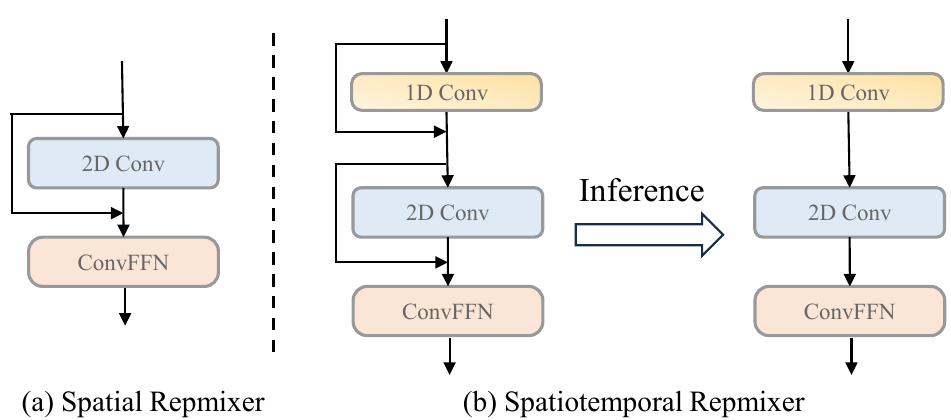}
  \caption{
\textbf{Spatiotemporal RepMixer.} (a) is a spatial RepMixer along with ConvFFN provided by MCi. (b) is the spatiotemporal RepMixer which adds 1D depthwise convolutional layer along with skip-connection and it can also be reparameterized at inference time. The norm layer and nolinearity are omitted for simplicity. 
}
\label{fig:conv1d}
\vspace{-4mm}
\end{figure}

\begin{figure}[!t]
  \includegraphics[width=0.5\textwidth]{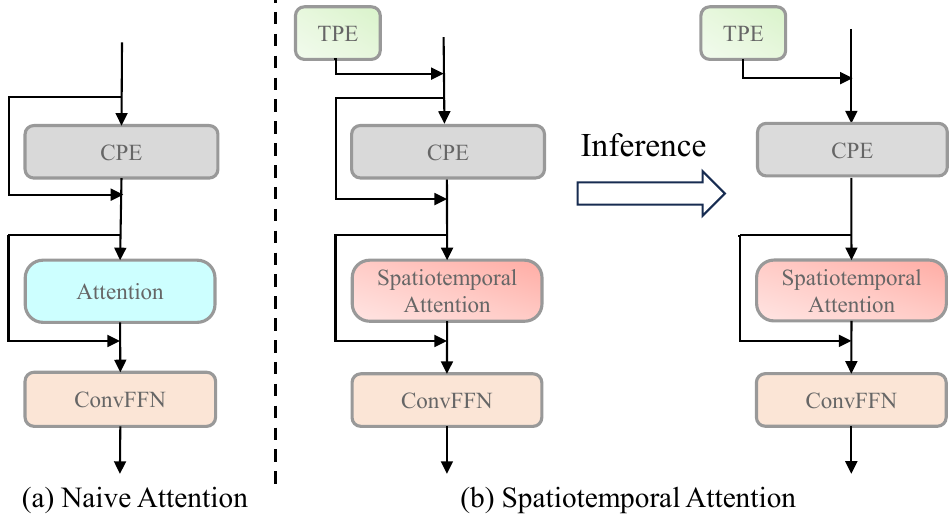}
  \caption{
\textbf{Spatiotemporal Attention.} (a) is a native attention block along with conditional positional embeddings and ConvFFN provided by MCi. In (b) we add learnable temporal positional embeddings (short for TPE) and update the native attention to spatiotemporal attention while maintaining other design elements. The norm layer and nolinearity are omitted for simplicity. 
}
\label{fig:attn3d}
\vspace{-4mm}
\end{figure}

\subsubsection{Block Design}
MCi is a hybrid vision transformer architecture that combines convolutional and transformer designed to effectively capture local and global information. However, as an image-text foundation model, it lacks the ability to understand temporal relationships. In this part, we present our block improvement to better understand the spatiotemporal relationships inside the video. 
\paragraph{\bf{Spatiotemporal Repmixer.}} 
 Inspired by~\cite{vittad,timesformer,r21d}, introducing a simple temporal modeling block can effectively enhance the model’s video understanding capabilities. Specifically, we construct a 1D depthwise convolutional layer to process temporal modeling before the 2D depthwise convolutional layer shown in Fig~\ref{fig:conv1d}. Specifically, for an input tensor X, the Spatiotemporal RepMixer can be implemented as:
 {\small
 \[X' = DWConv1D(BN(X)) + X,\]
 \[Y = DWConv2D(BN(X')) + X'.\]
 }
 where $X'$ needs to be rearranged to align with corresponding operations. It can also be reparameterized at inference time as:
 {\small
 \[X' = DWConv1D(X),\]
 \[Y = DWConv2D(X').\]
 } Such design will not affect the subsequent channel mixer ConvFFN~\cite{fastvit}.

\paragraph{\bf{Spatiotemporal Attention.}} 
Since the attention block in MCi is to learn global image information. Inspired by ViCLIP~\cite{internvid}, we update the native attention to spatiotemporal attention while maintaining other design elements including conditional positional encodings shown in Fig~\ref{fig:attn3d}. 
MCi adopts conditional positional encodings (CPE) generated by a depth-wise convolution operator but can only capture spatial relationships.
We add a learnable temporal position encoding (TPE) before CPE to provide the model with temporal position information. Specifically, for an input tensor $X$, the process of adding positional encoding can be described as follows:
{\small
 \[X' = X + TPE,\]
 \[Y = X' + CPE(X').\]
 }
During inference, we follow the structural reparameterization mentioned before:
{\small
\[X' = X + TPE,\]
\[Y = CPE(X').\]
}
In this way, we successfully add spatial and temporal position encoding to the input and send it into the subsequent Spatiotemporal Attention module for global spatiotemporal representation modeling.

\subsection{Training}
\paragraph{\bf{Settings.}} 
 For the video input of our MobileViCLIP, we sample 8 frames from the raw videos with an input resolution of $256\times256$. We train both MobileViCLIP-Tiny and MobileViCLIP-Small using AdamW~\cite{adamw} optimizer for 3 epochs with a learning rate of 1e-5, and we set the warm-up epoch to 0.6 with a linear schedule. During training, we use checkpoint technology to reduce the model's memory requirements at the expense of training speed, and then
 adopt 8 NVIDIA GeForce RTX 3090 GPUs, and the batch size is set to 64 for each GPU. When we fully fine-tune our model on downstream text-to-video retrieval datasets~\cite{msrvtt,didemo,anet-captions}, we also set batch size to 64 and adopt 10 epochs with a learning rate of 1e-5 and set the warm-up epoch to 1. Since our MobileViCLIP is sufficiently lightweight and requires less training data compared to other foundation models~\cite{internvideo2,internvl,unmaskedteacher}, which is significantly lower than other video-text foundation models.

 \paragraph{\bf{Pre-trained Dataset.}}
 Our baseline model MobileCLIP~\cite{mobileclip} adopted reinforced DataComp~\cite{datacomp} using its multi-modal dataset reinforcement strategy and created DataCompDR-1B by reinforcing DataComp-1B. Such a combination of both real and synthetic captions along with filtered high-quality images enables MobileCLIP to obtain the best zero-shot retrieval and classification performance. Based on the high-quality pretraining stage, we can directly fine-tune the model on the video-text dataset to directly learn video-text representations without designing multi-stage alignment and understanding. As our training dataset, we use InternVid-10M-FLT~\cite{internvid} which includes 10M high-quality YouTube videos spanning 16 scenes with about 6 thousand actions. The fine-grained video captions powered by large language models (LLM) can also ensure accurate descriptions of complex interactions inside these videos. The summary of datasets can be described in Table~\ref{table:dataset}, sufficient training corpus data ensures MobileViCLIP's generalization performance.

 \begin{table}[]
 \resizebox{0.47\textwidth}{!}{
\begin{tabular}{c|ccc}
\hline
Training Stage & Dataset           & Domain        & Scale          \\ \hline
Image-Text learning    & DataCompDR        & Web Image     & 1B   \\ \hline
Video-Text learning    & InternVid-10M-FLT & Youtube Video & 10M  \\ \hline
\end{tabular}
}
\caption{\textbf{Summary of datasets used in MobileViCLIP training}. DataCompDR is adopted by MobileCLIP, we fine-tune MobileViCLIP on InternVid-10M-FLT based on pre-trained MobileCLIP.}
\label{table:dataset}
\vspace{-5mm}
\end{table}

 \paragraph{\bf{Data Augmentation.}} 
We input MobileViCLIP with paired video and text $(v,t)$. During training, we randomly select 8 frames from the window, then resize and randomly crop the video frames to the resolution of $256\times256$, using bilinear interpolation. Then we horizontally flip the given frame randomly with a given probability 0.5. During inference, we only resize the frame to the resolution of $256\times256$ using bilinear interpolation.
\vspace{-5mm}
\paragraph{\bf{Training Objectives.}} 
We use simple video-text contrastive (VTC) loss to maximize the agreement between the paired video and text embeddings. Specifically, it minimizes InfoNCE loss using global video and text features, the video-text contrastive loss $L_{VTC}$ is given as:

\begin{align*}
L_{V2T}&=-\sum_{i=1}^{N}{\log\frac{\exp(sim(v_{i},t_{i})/\tau)}{\textstyle \sum_{j=1}^{N}\exp(sim(v_{i},t_{j})/\tau)}}, \\
L_{T2V}&=-\sum_{i=1}^{N}{\log\frac{\exp(sim(t_{i},v_{i})/\tau)}{\textstyle \sum_{j=1}^{N}\exp(sim(t_{i},v_{j})/\tau)}}, \\
L_{VTC} &= \frac{1}{2} (L_{V2T} + L_{T2V}),
\end{align*}
where $v$ and $t$ denote the learned video and text embeddings respectively. $sim()$ computes the cosine similarity between two features. $\tau$ is the learnable temperature. $N$ is the number of samples per batch.

\paragraph{\bf{Evaluation Datasets and Metrics.}} 
We used the following three text-to-video retrieval datasets to verify the video-text representation performance and explore the architecture design: MSR-VTT~\cite{msrvtt}, DiDeMo~\cite{didemo}, and ActivityNet-Captions~\cite{anet-captions}. MSRVTT is a widely used text-to-video retrieval dataset containing around 10,000 video clips across various scenes, each accompanied by 20 unique text descriptions. DiDeMo comprises over 10,000 videos, divided into 26,892 five-second video segments, each segment corresponding to multiple descriptions. ActivityNet is a large-scale dataset consisting of 200 categories, and most videos have durations between 5 and 10 minutes. We also test our MobileViCLIP on four widely-used action recognition datasets to verify the capacibility of zero-shot action classification: Kinetics-400~\cite{k400}, UCF-101~\cite{ucf101}, HMDB-51~\cite{hmdb51} and Something-Something V2~\cite{ssv2}. Kinetics-400 is a well-established dataset for action recognition, containing 400 different action categories with video clips around 10 seconds in length. UCF-101 includes 13,320 video clips categorized into 101 distinct actions. HMDB-51 consists of 6,766 short video clips divided into 51 unique action categories. Something-Something V2 is an extensive dataset comprising 220,847 videos, capturing a wide variety of actions involving daily object interactions. For Something-Something V2, We transform it into a multiple-choice task, namely SSV2-MC. Each video is paired with 173 descriptions, 172 of which are negative distractors. Models are required to retrieve the single correct description.
As our evaluation metric, we only report the Top-1 (Recall@1) text-to-video (T2V) and video-to-text (V2T) retrieval accuracy across these three text-to-video retrieval datasets~\cite{msrvtt,didemo,anet-captions}. We report Top-1 Accuracy (Acc) on these four action classification datasets~\cite{k400,ucf101,hmdb51,ssv2} under zero-shot settings. Specifically, we construct templates for these action categories, such as "a video of $action$", and verify the model's zero-shot classification capability by evaluating Top-1 V2T.

\section{Main Results}
\paragraph{\bf{Text-to-Video Retrieval.}}
We compare our MobileViCLIP with other models in the video-text retrieval task shown in Table~\ref{table:zs_vtr} and Table~\ref{table:ft_vtr}. We report R@1 scores for text-to-video (T2V) and video-to-text (V2T) tasks to evaluate retrieval performance. In addition, we also evaluate the model's throughput on GPU, parameters, FLOPs, and latency across various platforms, including Intel Xeon Gold 6248 CPU, V100 GPU, and iPad Air 2020 mobile device. The iPad  Air  2020 employs the same A14 Bionic SoC as the widely adopted iPhone 12 Pro Max\cite{mobileclip}—namely, a six‑core CPU, four‑core GPU, and 16‑core Neural Engine built on the 64‑bit ARMv8.5‑A architecture—with the only hardware distinction being memory capacity (4 GB versus 6 GB  LPDDR4X, respectively).
For iPad latency measurements, we export the models using Core ML Tools (v7.1) with iOS 18. Latency benchmarking batch size is set to 1 as in a real-world scenario and the optimal possible batch size to measure throughput before going out of memory. As shown in Table~\ref{table:zs_vtr} and Table~\ref{table:ft_vtr}, both versions of MobileViCLIP introduce negligible additional latency during inference while achieving superior performance. For zero-shot text-to-video retrieval, the performance of MobileViCLIP-Small is comparable to InternVideo2-L14 on MSR-VTT~\cite{msrvtt} (with +0.4/-0.6 T2V/V2T zero-shot R@1 scores) and outperforms InternVideo2-B14 on DiDeMo~\cite{didemo} (with +0.4/+2.0 T2V/V2T zero-shot R@1 scores). Except for long video datasets which consist of several minutes long videos like ActivityNet~\cite{anet-captions}, MobileViCLIP-Small still outperforms InternVideo2-S14 (with +0.8/+8.3 T2V/V2T zero-shot R@1 scores). 
For fine-tuned text-to-video retrieval, though our MobileViCLIP has a lower upper bound than the video clip model with larger parameters, our MobileViCLIP-Small still achieves a comparable performance with ViCLIP~\cite{internvid}. 
When we evaluate the efficiency-related metrics of these models, we find that our MobileViCLIP is more efficient than existing video-text models. With a similar number of parameters to InternVideo2-S14, our MobileViCLIP-Small takes only half the amount of FLOPs and is $6.75\times$ faster than it on mobile devices. However, due to the optimization of transformers on GPUs, MobileViCLIP performs slightly worse than InternVideo2-S14 on GPUs. When compared to other larger-scale video-text models, MobileViCLIP has a significant advantage in these metrics.
In summary, our MobileViCLIP shows reliable video text retrieval performance, and its structure based on an efficient image-text understanding model gives it significant advantages over other video CLIPs in terms of latency, FLOP, Params, and throughput.  
\begin{table*}[]
\centering
\resizebox{1.0\textwidth}{!}{
\begin{tabular}{cccccccccccccc}
\hline
\multirow{2}{*}{Params Range} &\multirow{2}{*}{Method} & \multirow{2}{*}{GPU Throughput} & \multirow{2}{*}{Params(M)} & \multirow{2}{*}{FLOPs(G)} & \multicolumn{3}{c}{Latency(ms)} & \multicolumn{2}{c}{MSR-VTT} & \multicolumn{2}{c}{DiDeMo} & \multicolumn{2}{c}{ActivityNet} \\
\cmidrule(r){6-8} \cmidrule(r){9-10}  \cmidrule(r){11-12} \cmidrule(r){13-14}
&&   videos/s    &    video+text     &     video+text      &    CPU(video+text)    &   GPU(video+text)   & Mobile(video+text) & T2V     & V2T & T2V    & V2T & T2V         & V2T \\ \hline
\multirow{3}{*}{$<$55M}&MobileCLIP-S0~\cite{mobileclip}&     114.8      &     11.4+42.4      &     14.8+1.3      &   71.3+7.2      &  12.1+5.3   &    9.9+2.6    &     33.2    &  28.6   &   31.2     &  31.5   &    23.8    &   22.4  \\
&MobileCLIP-S0(10M)&  114.8  &     11.4+42.4      &     14.8+1.3      &    71.3+7.2     &  12.1+5.3   &    9.9+2.6    &     37.2    &   36.9  & 33.5       &  33.6  &  26.0    &   26.5 \\
&\textbf{MobileViCLIP-Tiny}&  89.5  &     11.5+42.4      &     14.9+1.3      &    85.5+7.2     &  15.4+5.3   &    12.4+2.6    &    \textbf{38.7}     &  \textbf{38.1}   &  \textbf{37.1}      &  \textbf{37.0}  &     \textbf{29.3}        &  \textbf{28.2}   \\ \hline
\multirow{5}{*}{55-135M}&VideoCLIP~\cite{videoclip}&    321.8    &     8.4+110      &     9.0+11.3      &     42.9+29.1    &  14.5+9.5   &   4.5+7.2     &    10.4     & -    &     16.6   &  -   &    -         &   -  \\ 
&MobileCLIP-S2~\cite{mobileclip}&     66.1      &     35.8+63.4      &     48.4+4.3      &  178.5+15.4       &  20.1+8.1   &    24.7+4.0    &  38.8       &   32.8  &    35.5    &  35.3   &     29.6        &   27.7  \\
&InternVideo2-S14~\cite{internvideo2}&55.5 &     23.0+110      &     83.3+11.3      &    488.5+29.1     &   23.4+9.5  &    275+7.2    &    35.6     &  35.9      &  33.7  &     35.5  &       34.5      &  23.6   \\
&MobileCLIP-S2(10M)& 66.1   &     35.8+63.4      &     48.4+4.3      &     178.5+15.4    &   20.1+8.1  &   24.7+4.0    &    41.2    &  40.1  &  36.9      &  39.5  &     32.5        &  30.8   \\  
&\textbf{MobileViCLIP-Small}& 47.9   &     36.0+63.4      &     49.5+4.3      &     195.7+15.4    &   30.9+8.1  &   37.8+4.0    &    \textbf{42.5}    &  \textbf{43.5}  &  \textbf{40.7}      &  \textbf{41.1}  &     \textbf{35.3}        &  \textbf{31.9}   \\  \hline
\multirow{6}{*}{135-230M}&CLIP4CLIP~\cite{clip4clip}&   99.6     &      87.8+63.4     &     37.2+4.3      &    84.8+15.4     &  11.8+8.1   &    18.0+4.0    &    32.0     &  -   &   -     &   -  &  -           &  -   \\
&Open-VCLIP++(ViT-B/16)~\cite{open-vclip++}&   21.9     &      39.8+110     &     89.6+11.3      &    384.8+29.1     &  49.3+9.5   &   175.4+7.2     &    34.6     &  35.4   &    -    &  -   &   -     &  -   \\
&VINDLU~\cite{vindlu}&    27.8    &      121.2+110     &     185.6+11.3      &    775.2+29.1     &  38.9+9.5   &   712.5+7.2     &    32.0     &  -   &   36.9     &   -  &  30.9           &  -   \\
&UMT-B~\cite{unmaskedteacher}&    27.8    &      121.2+110     &     185.6+11.3      &    775.2+29.1     &  38.9+9.5   &   712.5+7.2     &    35.2     &  30.3   &   \textbf{41.2}     &   \textbf{40.8}  &  35.5          &  32.8   \\
&MobileCLIP-B~\cite{mobileclip}&      32.0     &    86.4+63.4       &      141.8+4.3     &    259.3+15.4     &  46.5+8.1   &    96.2+4.0    & \textbf{40.8}      &  37.8   &  36.2      & 37.4    & 34.4    & 31.9    \\
&InternVideo2-B14~\cite{internvideo2}&22.1 &      89.4+110     &     254.8+11.3      &    992.0+29.1     &   57.9+9.5  &    661+7.2    &    40.3     &  \textbf{48.5}   &    40.3    &   39.1  &   \textbf{41.5}    &   \textbf{38.8}  \\ \hline
\multirow{6}{*}{$>$230M}&CLIP~\cite{clip}&       6.9      &      304+124     &      649+9.7     &     1865.5+25.9    &   108.1+9.5  &    1875.5+ 7.0   &    30.4     &  24.2   &    12.7    &   18.7  &     9.1        &  13.2   \\
&Open-VCLIP++(ViT-L/14)~\cite{open-vclip++}& 6.9   &      304+124     &    649+9.7       &    1865.5+25.9     &   108.1+9.5  &   1875.5+7.0    &   39.0      &  37.8   &  -     & -    &    -      &  -   \\
&InternVideo-L~\cite{internvideo}&  5.9  &      308.9+124     &    830.7+9.7       &    2887.1+25.9     &   174.2+9.5  &   2298.7+7.1    &    40.7     &  39.6   &    31.5   &  33.5   &      30.7      &   31.4  \\
&ViCLIP~\cite{internvid}&     5.9    &      308.9+124     &     830.7+9.7      &    2887.1+25.9     &  174.2+9.5   &   2298.7+7.0     &    \textbf{42.4}     &  41.3   &   38.7     &   39.1  &    32.1         &   31.4  \\
&UMT-L~\cite{unmaskedteacher}&    4.7    &    429.4+335.1       &      652.2+42     &   2672.4+73.1      &  196.1+18.1   &    2373.7+20.5    &    40.7     &  37.1   &   \textbf{48.6}     &   \textbf{49.9}  &  41.9      &  39.4   \\
&InternVideo2-L14~\cite{internvideo2}&4.9 &      308.9+335.1     &     830.7+42      &    2887.1+73.1     &  178.1+18.1   &   2298.7+20.5     &     42.1    &   \textbf{44.1}  &     42.8   &  43.2   &   \textbf{43.6}     &  \textbf{40.7}   \\
\hline
\end{tabular}
}
\caption{\textbf{Results of zero-shot text-video retrieval on MSR-VTT~\cite{msrvtt}, DiDeMo~\cite{didemo} and ActivityNet~\cite{anet-captions}.} T2V and V2T are short for R@1 scores for text-to-video and video-to-text tasks. ``video" means the efficient metrics related to video branch in the model, while ``text" means the efficient metrics related to text branch in the model. ``M" means $10^6$. ``G" means $10^9$. ``ms" means millisecond. MobileCLIP(10M) is our baseline fine-tuned on InternVid-10M-FLT~\cite{internvid}.}
\label{table:zs_vtr}
\end{table*}
\begin{table*}[]
\centering
\resizebox{1.0\textwidth}{!}{
\begin{tabular}{ccccccccccccccc}
\hline
\multirow{2}{*}{Params Range}&\multirow{2}{*}{Method}&\multirow{2}{*}{GPU Throughput} & \multirow{2}{*}{Params(M)} & \multirow{2}{*}{FLOPs(G)} & \multicolumn{3}{c}{Latency(ms)} & \multicolumn{2}{c}{MSR-VTT} & \multicolumn{2}{c}{DiDeMo} & \multicolumn{2}{c}{ActivityNet} \\
\cmidrule(r){6-8} \cmidrule(r){9-10}  \cmidrule(r){11-12} \cmidrule(r){13-14}
 & &videos/s &   video+text     &    video+text     &    CPU(video+text)    &  GPU(video+text)   & Mobile(video+text) & T2V     & V2T & T2V    & V2T & T2V         & V2T \\ \hline
\multirow{2}{*}{{$<$55M}}&MobileCLIP-S0(10M)&  114.8  &     11.4+42.4      &     14.8+1.3      &    71.3+7.2     &  12.1+5.3   &    9.9+2.6    &    43.9     &   43.1  &   42.8    &  41.7  & 40.1   &  41.2 \\
&\textbf{MobileViCLIP-Tiny}& 89.5    &     11.4+42.4      &     14.9+1.3      &    85.5+7.2     &  15.4+5.3   &    12.4+2.6    &    \textbf{45.1}     &  \textbf{44.9}   &     \textbf{44.2}   & \textbf{44.7}    &       \textbf{42.7}      &   \textbf{43.3}  \\  \hline
\multirow{3}{*}{{55-135M}}&All-in-one~\cite{allinone} &   30.2    &     86.2     &     141.0      &   248.4      &  27.2   &    102.5    &   37.9      &   -  &   32.7     &  -   &      22.4       &   -     \\
&VideoCLIP~\cite{videoclip} & 321.8      &      8.4+110     &      9.0+11.3     &    42.9+29.1     &  14.5+9.5   &   4.5+7.2     &     30.9    &   -  &   -     &   -  &      -       &   -  \\
&\textbf{MobileViCLIP-Small}& 47.9   &     36.0+63.4      &    49.5+4.3       &      195.7+15.4   &   30.9+8.1  &   37.8+4.0     &    \textbf{49.3}     &   \textbf{50.3}  &   \textbf{48.2}     &  \textbf{48.5}   &    \textbf{47.3}   &  \textbf{46.4}   \\ \hline
\multirow{6}{*}{{135-230M}}&VIOLET~\cite{violet} &    29.8   &      87.6+110     &      65.6+11.3     &     290.3+29.1    &   37.6+9.5  &    -    &    34.5     &  -   &    32.6    &  -   &      -       &  -    \\
&Frozen~\cite{frozen}&  26.3         &     85.8+66.3      &     254.8+3.3      &     652.1+25.7    &   41.8+4.7  &    543.8+7.1    &   31.0      &  -   &    34.6    &   -  &      -       &   -  \\
&CLIP4CLIP~\cite{clip4clip} & 99.6       &      87.8+63.4     &     37.2+4.3      &    84.8+15.4     &  11.8+8.1   &     18.0+4.0   &    45.6     &  45.9   &   43.0     &  43.6   &       40.3      &   41.6  \\
&ALPRO~\cite{alpro} & 27.1      &      121+110     &      196+11.3     &    324.2+29.1     &   40.2+9.5  &    1124.9+7.2    &   33.9      &  -   &    35.9    &   -  &     -        &   -  \\
&VINDLU~\cite{vindlu} &    27.8       &     121.2+110      &      185.6+11.3     &    775.2+29.1     &  38.9+9.5   &   712.5+7.2    &      46.5   &  -   &  61.2    & -   &  55.0   &  -  \\ 
&UMT-B~\cite{unmaskedteacher} &    27.8       &     121.2+110      &      185.6+11.3     &    775.2+29.1     &  38.9+9.5   &   712.5+7.2      &   \textbf{51.0}      &   \textbf{49.0}  &  \textbf{61.6}    & \textbf{ 59.5}  &   \textbf{58.3}    &  \textbf{56.0}  \\ 
\hline
\multirow{4}{*}{{$>$230M}}
&CLIP~\cite{clip}  &    6.9       &     304+124     &      649+9.7     &     1865.5+25.9    &  108.1+9.5   &     1875.5+7.0   &    38.2     &  38.7   &  32.2      &  33.9   &       26.1      &  26.9   \\
&ViCLIP~\cite{internvid} & 5.9          &     308.9+124      &     830.7+9.7      &    2887.1+25.9     &   174.2+9.5  &   2298.7+7.0     &    52.5     &   51.8  &  49.4      &  50.2   &      49.8       &   48.1  \\
&VINDLU-L~\cite{vindlu} &  4.7         &   429.4+335.1        &      652.2+42     &    2672.4+73.1     &  196.1+18.1   &   2373.7+20.5    &    48.8     &  -   &  59.8    &  -  &    55.9     & -   \\
&UMT-L~\cite{unmaskedteacher} &  4.7         &   429.4+335.1        &      652.2+42     &    2672.4+73.1     &  196.1+18.1   &   2373.7+20.5    &  \textbf{58.8}     & \textbf{58.6} &  \textbf{70.4}   &  \textbf{65.7 }   &  \textbf{66.8}  &   \textbf{64.4 }\\
\hline
\end{tabular}
}
\caption{\textbf{Results of fine-tuned text-video retrieval on MSR-VTT~\cite{msrvtt}, DiDeMo~\cite{didemo}, and ActivityNet~\cite{anet-captions}
}. T2V and V2T are short for R@1 scores for text-to-video and video-to-text tasks. ``video" means the efficient metrics related to video branch in the model, while ``text" means the efficient metrics related to text branch in the model. ``M" means $10^6$. ``G" means $10^9$. ``ms" means millisecond. MobileCLIP(10M) is our baseline fine-tuned on InternVid-10M-FLT~\cite{internvid}.}
\label{table:ft_vtr}
\end{table*}

\vspace{-4mm}
\paragraph{\bf{Zero-Shot Action Recognition.}}
We further explore the action recognition capability of MobileViCLIP under zero-shot settings. We randomly construct complete description sentences for each action following the prompts provided by~\cite{internvideo2} and use R@1 scores for video-to-text task to evaluate the top-1 accuracy provided by MobileViCLIP. As shown in Table~\ref{table:zs_ar}, MobileViCLIP-Small can achieve action recognition capabilities better than InternVideo-S14 with an improvement of +1.0 Top-1 accuracy on Kinetics-400~\cite{k400}, +2.0 Top-1 accuracy on UCF-101, +1.2 Top-1 on HMDB-51, but -0.3 Top-1 accuracy on Something-Something V2. 
We also find that MobileViCLIP performs well on the HMDB-51~\cite{hmdb51} dataset which is even better than InternVideo2-L14. 
Although MobileViCLIP is fine-tuned on a video-text retrieval dataset, the rich video-text pairs it has learned still enable MobileViCLIP to achieve strong zero-shot action recognition capabilities. 
\begin{table}[]
\resizebox{0.47\textwidth}{!}{
\begin{tabular}{ccccc}
\hline
Method          & K400 & UCF-101 & HMDB-51 & SSV2-MC \\ \hline
MobileCLIP-S0~\cite{mobileclip}   &   48.6   &  67.9   &  41.2    & 33.2        \\
MobileCLIP-S0(10M)   &   53.0   &   73.3  &   46.0  &  35.5       \\ 
\textbf{MobileViCLIP-Tiny}   &   \textbf{56.2}   &   \textbf{75.4}  &   \textbf{48.8}   &  \textbf{39.3}       \\  \hline
MobileCLIP-S2~\cite{mobileclip}   &   57.0   &  73.9   &  45.7    &  41.8      \\
MobileCLIP-S2(10M)   &   60.4   &  77.1   &  51.8    &  44.1      \\
InternVideo2-S14~\cite{internvideo2} &   62.1   & 79.0    &   49.2   &   \textbf{46.4}      \\
\textbf{MobileViCLIP-Small}   &  \textbf{63.1}    &  \textbf{81.0}   &   \textbf{53.7}   &     46.1   \\   \hline
MobileCLIP-B~\cite{mobileclip}    &   61.1   &  77.7   &  50.1    &  44.9       \\
InternVideo2-B14~\cite{internvideo2} &   \textbf{67.7}   &  \textbf{83.4}   &   \textbf{52.5}   &   \textbf{55.9}      \\  
\hline
CLIP~\cite{clip}            &   58.4   &   68.9  &  43.2    & 29.6        \\
InternVIdeo2-L14~\cite{internvideo2} &   \textbf{70.7}   &  \textbf{85.9}   &   \textbf{53.2}  &    \textbf{59.6}     \\ 
\hline
\end{tabular}
}
\caption{\textbf{Results of zero-shot action recognition}. We report top-1 accuracy on Kinetics-400~\cite{k400}, UCF-101~\cite{ucf101}, HMDB-51~\cite{hmdb51}, and Something-Something V2~\cite{ssv2}. MC is short for ``multiple choice". MobileCLIP(10M) is our baseline fine-tuned on InternVid-10M-FLT~\cite{internvid}. 
}
\label{table:zs_ar}
\end{table}

\vspace{-2mm}
\section{Ablation Study}

\paragraph{\bf{Effectiveness of Block Design in MobileViCLIP.}}
In this section, we ablate the block design in MobileViCLIP. For a fair comparison, we ablate the following experiments based on MSR-VTT for its sufficient training data and diverse text descriptions and then evaluate these variants based on R@1 scores for text-to-video on the test set of this dataset. Shown in Table~\ref{table:ablation_conv1d_attn3d}, both Spatiotemporal RepMixer and Spatiotemporal Attention can achieve higher results. We can also notice that the temporal positional encodings are important to help the model better understand the temporal changes of video. So we accept both of them in our MobileViCLIP.

\begin{table}[]
\resizebox{0.47\textwidth}{!}{
\begin{tabular}{cc}
\cline{1-2}
Model                                               & MSR-VTT  \\ \cline{1-2}
Baseline                                                 &   38.4   \\
+ Spatiotemporal Repmixer                                             &   39.5   \\
+ Spatiotemporal Attention w/o TPE               & 39.1 \\
+ Spatiotemporal Attention w/ TPE          &          39.6\\
+ Spatiotemporal Repmixer + Spatiotemporal Attention w/ TPE &      \textbf{40.1} \\ \cline{1-2}
\end{tabular}
}
\caption{\textbf{Ablation of block designs}. Baseline is MobileCLIP fine-tuned on MSR-VTT. TPE is short for temporal positional encodings as mentioned before.}
\label{table:ablation_conv1d_attn3d}
\end{table}

\vspace{-6mm}
\paragraph{\bf{Freeze Text Branch or Not.}}
We further ablate on whether to freeze the text branch during training because some works~\cite{vifi-clip} train both vision and text branches simultaneously, while other works~\cite{internvideo2} freeze the text branch. Shown in Table~\ref{table:temporal_modeling}, we find that activating the text branch brings no help to the final video-to-text retrieval results and brings additional training consumption. So we freeze the text branch.

\begin{table}[]
\resizebox{0.47\textwidth}{!}{
\begin{tabular}{cccc}
\hline
Video Branch    &Text Branch    & Training Memory  & MSR-VTT \\ \hline
Activate         &Freeze         &   24G      &    \textbf{40.1}     \\
Activate         &Activate   &    27G        &     39.7    \\ \hline
\end{tabular}
}
\caption{\textbf{Freeze branch or not}. In order to directly reflect the consumption of training memory, we do not use the checkpoint technology to reduce the consumption of video memory here. We use V100 GPU to do this ablation experiment.}
\label{table:temporal_modeling}
\end{table}

\vspace{-5mm}
\paragraph{\bf{Validate Video-Text Representation Capability for Video Downstream Tasks.}} 
To verify the generalization performance of our MobileViCLIP in other video downstream tasks, we report the results on following video downstream tasks including video temporal grounding, zero-shot temporal action detection and video captioning. Video temporal grounding is to identify specific moments or highlights from a video corresponding to textual descriptions. It relies on excellent video-text pair representation. To validate the video-text representation of our MobileViCLIP, we treat MobileViCLIP as the feature extractor for QVHighlights~\cite{qvhighlights} and use CG-DETR~\cite{cg-detr} as the temporal grounding detector. For a fair comparison, we also finetune the MobileCLIP-S0 with InternVid-10M-FLT and use it as the feature extractor to let MobileCLIP learn video-text representation. As shown in Table~\ref{table:test_mr}, MobileViCLIP-Small shows stronger generalization performance on both moment retrieval and highlight detection tasks, even better than the combination of CLIP~\cite{clip} and SlowFast~\cite{slowfast} features which are adopted by many temporal grounding models~\cite{cg-detr,umt}. Zero-shot temporal action detection seeks to identify and locate actions in untrimmed videos when the action label is unseen. As shown in Table~\ref{table:tad}, our MobileViCLIP-Small performs slightly better than ViT-L/14 on T3AL~\cite{t3al}. We also evaluate our MobileViCLIP on video captioning task on CLIP4Caption~\cite{clip4caption} and find that MobileViCLIP-Small performs better than VIT-B/32. 

\begin{table}[]
\resizebox{0.47\textwidth}{!}{
\begin{tabular}{cccccccc}
\hline
 \multirow{3}{*}{Backbone}             & \multicolumn{5}{c}{Moment Retrieval}     & \multicolumn{2}{c}{Highlight Detection}      \\ \cline{2-8} 
              & \multicolumn{2}{c}{R1}& \multicolumn{3}{c}{mAP}   & \multicolumn{2}{c}{\textgreater{}=Very Good}       \\ \cline{2-8} 
          & @0.5 & @0.7 & 0.5 & 0.75 & Avg & mAP                      & HIT@1 \\ \hline
CLIP~\cite{clip}  &  66.8    &  49.1   &   65.8   &  44.1   &     43.5     &   39.3 & 64.0     \\
CLIP~\cite{clip}+SlowFast~\cite{slowfast}  &  67.4    &   52.1     &  65.6   &  45.7 &  44.9   &  40.8  &  66.7  \\
MobileCLIP-S0~\cite{mobileclip}      &  66.0    &   48.0   &  64.9   &   42.4   &  41.8   &    39.3                      &   63.6    \\
MobileCLIP-S0 (10M) &   67.3   &  49.2    &  65.9   &   44.4   &  43.8   &  39.8                        &   64.8    \\  \hline
MobileViCLIP-Tiny &  68.4    &  51.7    &  66.7   &   45.3   &  44.9   &   40.6                       &   65.4    \\ 
MobileViCLIP-Small &  \textbf{68.9}    &   \textbf{53.3}   &  \textbf{67.8}   &  \textbf{47.3}    &  \textbf{46.3}   &        \textbf{41.3}   &   \textbf{67.5}    \\ \hline
\end{tabular}
}
\caption{\textbf{Performance comparison on QVHighlights~\cite{qvhighlights} validation splits with features from different backbones for CG-DETR~\cite{cg-detr}}. We calculate the average mAP score with IoU thresholds
ranging from 0.5 to 0.95 in 0.05 intervals.}
\label{table:test_mr}
\end{table}
\vspace{-2mm}
\begin{table}[]
\resizebox{0.47\textwidth}{!}{
\begin{tabular}{c|c|cccccc}
\hline
Method       & Backbone           & 0.3  & 0.4  & 0.5  & 0.6 & 0.7 & Avg.mAP \\ 
\hline
\multirow{2}{*}{T3AL$_{T=0}$} & Coca(ViT-L/14)     & 11.4 & 6.8  & 3.5  & 1.7 & 0.6 & 4.8  \\
& MobileViCLIP-Small & \textbf{12.5} & \textbf{7.4} & \textbf{3.6}  & \textbf{1.9} & \textbf{1.2} & \textbf{5.3}  \\ \hline
\end{tabular}
}
\vspace{-3mm}
\caption{
\textbf{Performance comparison on THUMOS14~\cite{thumos} with features from different backbones for T3AL~\cite{t3al}.} We evaluate the average mAP score with IoU thresholds ranging from 0.3 to 0.7 in 0.1 intervals.
}
\vspace{-3mm}
\label{table:tad}
\end{table}

\begin{table}[]
\resizebox{0.47\textwidth}{!}{
\begin{tabular}{c|c|cccc}
\hline
Method      & Backbone     & BLEU-4 & ROUGE-L & METEOR & CIDEr  \\ \hline
\multirow{2}{*}{CLIP4Caption} & ViT-B/32     & 46.1   & 63.7    & 30.7   & 57.7 \\
             & MobileViCLIP-Small &  \textbf{48.9}     &   \textbf{66.7}      &  \textbf{32.1}      &  \textbf{61.5}     \\ \hline
\end{tabular}
}
\caption{
\textbf{Performance comparison on MSR-VTT~\cite{msrvtt} with features from different backbones for CLIP4Caption~\cite{clip4caption}.} We report the standard captioning metrics, BLEU4~\cite{bleu}, ROUGE-L~\cite{rouge}, METEOR~\cite{meteor}, CIDEr~\cite{cider} of  CLIP4Caption. 
}
\label{table:videocaption}
\end{table}

\section{Latency Analysis}
As shown in Table~\ref{table:zs_vtr} and~\ref{table:ft_vtr}, 
the latency of the ViT model~\cite{internvideo2,unmaskedteacher,internvid,vindlu} is tens or hundreds of times that of the efficient models~\cite{mobileclip}. 
To delve deeper into the reasons, we decompose the convolution and attention modules for further exploration.  
As shown in Table~\ref{table:latency_GPU_latency}, we use ``Conv" to represent our Spatiotemporal RepMixer and ``Attn" to represent Spatiotemporal Attention. 
We also compare the channel mapping layer ``ConvFFN" in our MobileViCLIP which is also designed in FastViT~\cite{fastvit} and ``FFN" adopted in all ViT models as the channel mixer.
After that, we also simulate various usage scenarios of the input tensor by trying different input sizes and channel dimensions. 
We assume that the original input is 8 frames, and the patch size is $14\times14$ (ViT-L/14~\cite{internvideo} adopts it), $16\times16$ (ViT-B/16~\cite{internvideo} adopts it), $32\times32$ (CLIP4CLIP~\cite{clip} adopts it). For channel dimension, we try 64, 128, 256, 384, 512, 768, and 1024, which are adopted by many models~\cite{internvideo2,fastvit,repnext,repvit,repvgg}.  
Among these experiments, we can draw the following conclusions:

\vspace{-4mm}
\paragraph{\textbf{The Latency of Mobile Device is Much Bigger than GPU's under Fair Conditions.}} As shown in Table~\ref{table:latency_GPU_latency}, we observe that the latency tested on mobile devices is significantly higher than GPU's when the input size and channel dimension are the same. The gap is around 10 times under fair conditions(compare 0.343ms with 3.97ms) and the gap increases when the spatial size is bigger.  
This indicates that deploying the model on mobile devices requires considering the limited computational power of mobile devices, thus imposing higher requirements on the model.

\vspace{-4mm}
\paragraph{\textbf{The Latency is Positively Correlated with the Input Size.}}
When the input has a larger feature dimension or spatiotemporal size, the latency increases on both mobile devices and GPU shown in Table~\ref{table:latency_GPU_latency}. We also notice that the latency of attention on mobile device is consistent with the computational complexity $O(H^{2}W^{2}C)$ without considering $T$ (from 3.97ms to 60.1ms when width and height are doubled), but GPU latency does not match such complexity. This shows that the GPU has optimizations~\cite{flashattention} for attention mechanism operations, while the mobile device does not have. Controlling the size of the input tensor is also an effective measure to reduce latency on mobile device. 
\vspace{-4mm}
\paragraph{\textbf{Stacking more Attention Layers Brings Bigger Latency.}}
Table~\ref{table:latency_GPU_latency} shows that the attention mechanism incurs greater latency than convolution. When we construct more layers for each module shown in Table~\ref{table:latency_layers}, we can find that building more convolution layers has little effect on latency, but the latency introduced by the attention layers increases dramatically. Specifically, the latency of attention increases exponentially with the number of stacked layers(from 24.7ms to 296.4ms when stacking 10 layers) on mobile device. This indicates that mobile devices do not have GPU optimizations for attention mechanisms.

\begin{table}[]
\centering
\resizebox{0.47\textwidth}{!}{
\begin{tabular}{ccccccccc}
\hline
\multirow{2}{*}{\textbf{ST Size}} & \multirow{2}{*}{\textbf{Dim}} & \multicolumn{3}{c}{\textbf{GPU Latency (ms)}} & \multicolumn{3}{c}{\textbf{Mobile Latency (ms)}} \\ \cline{3-5} \cline{6-8}
& & \textbf{Conv} & \textbf{Attn} & \textbf{ConvFFN} & \textbf{Conv} & \textbf{Attn} & \textbf{ConvFFN} \\
\hline
\multirow{7}{*}{$8\times7\times7$}  & 64  & 0.066 & 0.070 & 0.191 & 0.11  & 0.21 & 0.33 \\
    & 128 & 0.067 & 0.095 & 0.198 & 0.16  & 0.76 & 0.57 \\
    & 256 & 0.070 & 0.132 & 0.209 & 0.26  & 1.17 & 0.80 \\
    & 384 & 0.071 & 0.198 & 0.214 & 0.37  & 1.48 & 1.01 \\
    & 512 & 0.073 & 0.237 & 0.225 & 0.49  & 2.01 & 1.22 \\
    & 768 & 0.075 & 0.275 & 0.236 & 1.32  & 2.98 & 1.75 \\
    & 1024 & 0.078 & 0.343 & 0.295 & 1.59  & 3.97 & 2.56 \\ 
\hline
\multirow{7}{*}{$8\times14\times14$}     & 64  & 0.071 & 0.228 & 0.196 & 0.13  & 2.48 & 0.69 \\
      & 128 & 0.072 & 0.295 & 0.205 & 0.47  & 5.01 & 0.95 \\
      & 256 & 0.072 & 0.452 & 0.209 & 1.14  & 12.2 & 1.42 \\
      & 384 & 0.074 & 0.762 & 0.220 & 1.55  & 18.6 & 1.98 \\
      & 512 & 0.076 & 1.052 & 0.227 & 2.01  & 24.7 & 2.66 \\
      & 768 & 0.079 & 1.608 & 0.241 & 2.66  & 44.6 & 4.37 \\
      & 1024 & 0.082 & 2.168 & 0.402 & 3.08  & 60.1 & 5.82  \\
\hline
\multirow{7}{*}{$8\times16\times16$}  & 64  & 0.070 & 0.691 & 0.228 & 0.15  & 3.45 & 0.83 \\
    & 128 & 0.071 & 0.766 & 0.236 & 0.85  & 8.76 & 1.27 \\
    & 256 & 0.073 & 0.974 & 0.249 & 1.25  & 18.3 & 1.91 \\
    & 384 & 0.074 & 1.511 & 0.262 & 1.71  & 28.1 & 2.65 \\
    & 512 & 0.077 & 2.026 & 0.284 & 2.16  & 47.3 & 3.46 \\
    & 768 & 0.081 & 3.071 & 0.325 & 3.03  & 70.4 & 4.85 \\
    & 1024 & 0.084 & 4.164 & 0.442 & 3.83  & 93.5 & 6.52 \\ 
\hline
\end{tabular}
}
\caption{\textbf{Comparison of GPU and Mobile Latency for Different Layers and Input Sizes.} ``ST" is short for spatiotemporal. We report the 1-layer component's GPU latency by averaging the latency caused by the stacked 30-layer network following the settings in MobileOne~\cite{mobileone}. For the mobile latency of each module, we only test a single layer.}
\label{table:latency_GPU_latency}
\end{table}

\begin{table}[]
\centering
\begin{tabular}{ccccc}
\hline
\multirow{2}{*}{\textbf{Layers}} & \multicolumn{4}{c}{\textbf{Mobile Latency (ms)}} \\ \cline{2-5}
& \textbf{Conv} & \textbf{Attn} & \textbf{FFN} & \textbf{ConvFFN} \\ \hline
1   & 2.01& 24.7 & 1.95 & 3.65 \\
10  & 2.98& 296.4 & 9.22 & 11.19 \\
20  & 3.18 & 543.3 & 17.79 & 22.39\\
30  & 3.47 & 783.8 & 26.78 & 32.33\\ \hline
\end{tabular}
\caption{\textbf{The impact of the number of layers on the latency of Spatiotemporal RepMixer, Attention layer, FFN, and ConvFFN.} The experiments are conducted with a dimension of 512 and an input size of $8\times14\times14$. 
}
\label{table:latency_layers}
\vspace{-4mm}
\end{table}

\section{Conclusion}
We have proposed an efficient video-text model MobileViCLIP for mobile devices. Equipped with minor changes inside the spatial modeling block of the efficient image-text model, our MobileViCLIP shows better spatiotemporal representation understanding capabilities compared to the baseline. MobileViCLIP is suitable for general tasks and yields good performance on several video-text retrieval and action recognition datasets. Furthermore, we comprehensively compare existing video-text models and then analyze reasons that will cause more latency. We hope it will serve as a guideline for future research.

\noindent {\small  \textbf{Acknowledgments.}  
This work is partially supported by the National Key Research and Development Program of China (No. 2022ZD0160900), Jiangsu Frontier Technology Research and Development Program (No. BF2024076), and Collaborative Innovation Center of Novel Software Technology and Industrialization.
}

{
    \small
    \bibliographystyle{ieeenat_fullname}
    \bibliography{main}
}

\end{document}